# Improving Training Result of Partially Observable Markov Decision Process by Filtering Beliefs

**Oscar LiJen Hsu**

**Abstract**

In this study I proposed a filtering beliefs method for improving performance of Partially Observable Markov Decision Processes(POMDPs), which is a method wildly used in autonomous robot and many other domains concerning control policy. My method search and compare every similar belief pair. Because a similar belief have insignificant influence on control policy, the belief is filtered out for reducing training time. The empirical results show that the proposed method outperforms the point-based approximate POMDPs in terms of the quality of training results as well as the efficiency of the method.

## Introduction

A point-based POMDP method uses sampling beliefs' idea to generate a limited belief set. Based on approximate idea, by analyzing the result of the sampling beliefs, many of the beliefs are similar and another approximate method is emergence. In Figure 1, two sample beliefs from hallway2 problem are given for explaining this idea. There are ninety-two elements in a vector. The dots symbol represents the elements which smaller than 0.001. Only few elements are detailed in a vector, and the positions of those elements are 2, 19, 69, 70, 71, 72, 73, 92. The visualization of this two belief vectors are shown in Figure 2 and Figure 3.

$$v_1 = \begin{bmatrix} \vdots \\ 0.4794 \\ \vdots \\ 0.0195 \\ \vdots \\ 0 \\ 0 \\ 0 \\ 0 \\ 0.0195 \\ \vdots \\ 0.4794 \end{bmatrix}, v_2 = \begin{bmatrix} \vdots \\ 0.4795 \\ \vdots \\ 0.0196 \\ \vdots \\ 0 \\ 0 \\ 0 \\ 0 \\ 0.0196 \\ \vdots \\ 0.4795 \end{bmatrix}$$

Figure 1: two sample beliefs

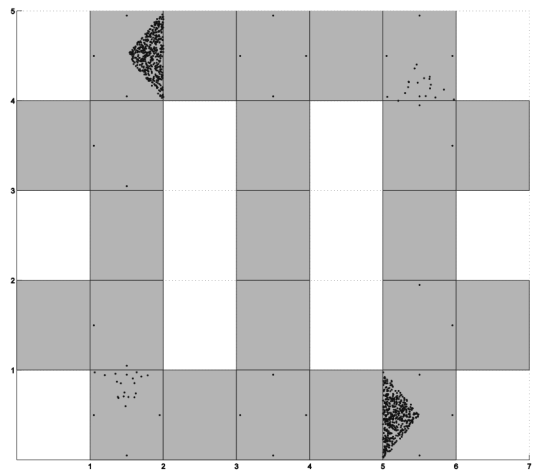

Figure 2: the visualization of the belief v1

A belief of Hallway2 problem is visualized as Figure 2. Each point represents one over one thousand probability and the number of points is decided by a belief. Points are randomly distributed in a triangle. Because there are ninety-two states in Hallway2 problem, there are ninety-two triangles in Figure 2 and each square includes four triangles which indicate four directions, up, right, down, left. Each triangle can be mapped to each element in a belief; therefore, the number of points is the number of probability. A triangle will be clearly shaped if the probability is large enough. A probability which less than one over one thousand is represented by one point. By the visualization of Figure 2, it is obvious that the belief intend to be located on two states which is two direction positions.

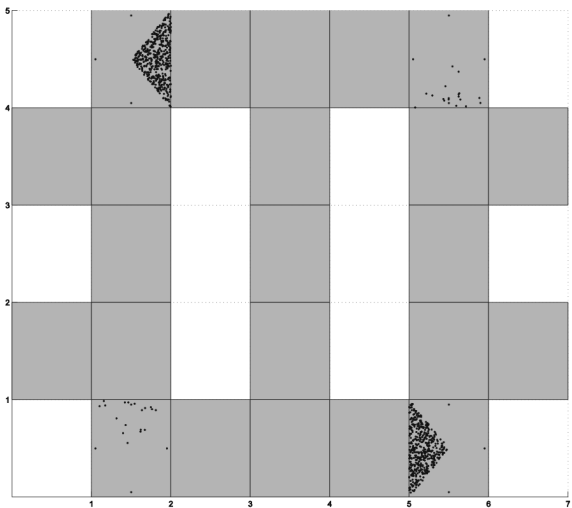

Figure 3: the visualization of the approximate Belief v2

There is another approximate belief Figure 3 which look almost like Figure 2. Based on that I like to solve Hallway2 problem by an approximate solution, the belief in Figure 3 can be filtered out for decreasing the time of finding solution.

Building on this insight, the goal of this study is to decrease computational time of value iteration in POMDP but with the same quality of control policy; in another words, this study increase quality of control policy but with the same computational time.

## Literature review

In this section, two important names are going to be detailed. They are the benchmark which is going to be computed, and the method of the control group which is going to be compared. The Hallway2 problem is a classical POMDP model which used to be called "large problem", it has been used in most POMDP research as a benchmark. A point-based approximate POMDP method, PERSEUS, is a good method for solving POMDP model and it provide Matlab program which is unsealed source code online. Most fundamental program in this study is modified from PERSEUS. Based on those, PERSEUS is chosen to be a control group in this study. However, this study is not that trying to prove a drawback of PERSEUS but that attempting to give a general filtering method which can be added into any point-based approximate POMDP method for improving training results.

## The Hallway2 problem

| 01<br>04,02<br>03 | 05<br>08,06<br>07 | 09<br>12,10<br>11 | 13<br>16,14<br>15 | 17<br>20,18<br>19 |       |
| --- | --- | --- | --- | --- | --- |
| 21<br>24,22<br>23 | 25<br>28,26<br>27 |       | 29<br>32,30<br>31 |       | 33<br>36,34<br>35 | 37<br>40,38<br>39 |
|       | 41<br>44,42<br>43 |       | 45<br>48,46<br>47 |       | 49<br>52,50<br>51 |       |
| 53<br>56,54<br>55 | 57<br>60,58<br>59 |       | 61<br>64,62<br>63 |       | 65<br>68,66<br>67 | <u>69</u><br><u>72</u>,70<br><u>71</u> |
|       | 73<br>76,74<br>75 | 77<br>80,78<br>79 | 81<br>84,82<br>83 | 85<br>88,86<br>87 | 89<br>92,90<br>91 |       |

Figure 4: The positions and the order

The basic idea of this model, is that there is an agent searching around an office. The region has been divided into twenty-three blocks. The states of Hallway2 are visualized as Figure 2. Each block includes four states that represent four different directions. The relationship between the order of states and positions with directions is as Figure 4. The numbers with underline are goal states, where are the objective to be achieved by the agent. The agent has five actions it can take, which include forward as Figure 5, turn-left as Figure 6, turn-right as Figure 7, turn-around as Figure 8, and stay.

Additionally, the agent is equipped four short range sensors to provide information about whether adjacent to walls. There are total seventeen observations of Hallway2 problem and these sensors give sixteen observations. The observations are combinations of four sides by four independent sensors which each have two possible probability value. 0.9 will be given by a sensor if there is an adjacent wall and the sensor senses the wall. 0.05 will be given by a sensor if there is no wall and the sensor senses a wall. The probabilities of observations of every states are given by multiplying four value from four sensors.

There are no sensors that give the current orientation of the agent, but there is an extra observation of the goal. This observation is only occurrence in the goal state and occurs deterministically. The goal state gives a reward of +1 while all other states give zero reward.(Littman et al., 1995)

**Algorithm 1** PERSEUS backup stage: $V_{n+1} \leftarrow \widetilde{H}_{PERSEUS} V_n$

**Input:** $V_n$
**Output:** $V_{n+1}$
1:    Set $V_{n+1} \leftarrow \emptyset$
2:    **Initialize** $\widetilde{B} \leftarrow B$
3:    **repeat**
4:       **Sample** a belief point b uniformly at random from $\widetilde{B}$
5:       **Compute** $\alpha \leftarrow$ backup(b)
6:       **if** $b \cdot \alpha \geq V_n(b)$ **then**
7:          **add** $\alpha$ to $V_{n+1}$
8:       **else**
9:          **add** $a' = arg \max_{\{\alpha_{n,i}\}_i} b \cdot \alpha_{n,i}$ to $V_{n+1}$
10:      **end if**
11:      **Compute** $\widetilde{B} \leftarrow \{b \in B : V_{n+1}(b) < V_n(b)\}$
12:   **until** $\widetilde{B} = \emptyset$

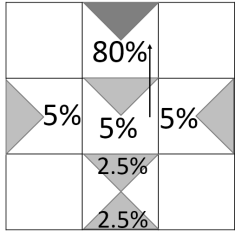

Figure 5 : Transition probabilities for a forward action

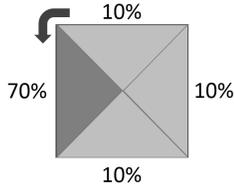

Figure 6 : Transition probabilities for the turn-left action

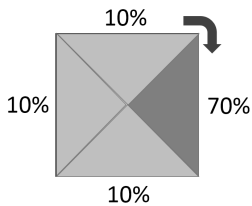

Figure 7 : Transition probabilities for a turn-right action

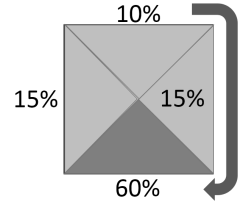

Figure 8 : Transition probabilities for a turn-around action

**A point-based approximate POMDP method, PERSEUS**

Spaan and Vlassis(2005) showed a modified method to obtain an approximate solution of POMDP. They used about ten thousand samples of belief vectors, the amount is difference between problems, to estimate the value function. The way they generated sample beliefs is to find the successor belief $b_{n+1}$ from current belief $b_n$ according to the Bayes' rule while the action, the next state and the observation are randomly chosen. An action is chosen according to a uniformly probability distribution; the next state is chosen according to the probability in the transition matrix of the action; and an observation is chosen according to the probability in the observation matrix of the action and the next state. The belief sampling process starts with finding a successor of the initial belief that could be a uniform belief. The process will keep finding a sequence of successor beliefs by finding a successor of the successor belief repeatedly until the total number of beliefs reaches the given required number. The approximate method of PERSEUS can improve the training time and make POMDP method more feasible. Because the backup stage of PERSEUS would be performed many times, more than one hundred rounds, to terminate, I can limit the training time or the number of repeat times as in practice, to get an acceptable result that approaches to the optimal. The evaluation results of PERSEUS shows, as compared with other method, a better control quality including terms of high expected reward and less training time.

Equations (1) (2) (3) (4) define backup projectors. (4) gives initial alpha vector in which each element is calculated from a minimum reward in reward parameters and discount factor $\gamma$. In (3), p(o|s',a) represents observation model and p(s'|s,a) represents transition model. In (2), $r_a$ represents a reward for a specific action a. b is a belief.

$$\text{backup(b)} = \alpha_{n+1}^b = arg \max_{\{q_a^b\}_{a \in A}} b \cdot q_a^b \quad (1)$$

$$q_a^b = r_a + \gamma \sum_o arg \max_{\{q_{a,o}^i\}_i} b \cdot q_{a,o}^i \quad (2)$$

$$q_{a,o}^i \equiv \begin{bmatrix} q_{a,o}^i(s1) \\ q_{a,o}^i(s2) \\ \vdots \end{bmatrix} \quad (3)$$

$$, q_{a,o}^i(s) = \sum_{s'} p(o|s',a) p(s'|s,a) \alpha_n^i(s')$$

$$\alpha_0 = \{\vec{v}\}, \forall x \in \vec{v}, x = \frac{\min(R)}{(1-\gamma)} \quad (4)$$

Algorithm 1 describes a point-based method to get optimal control policy. This algorithm is not performed one time to get solution. It is performed many times to improve solution at each time instead. At each time, this algorithm tries every belief in sample beliefs set. A backup projector $\alpha$ is computed for a belief. If a product of belief and $\alpha$ is bigger than maximum product of the same belief and a vector in $V_n$, this α will be preserved in output $V_{n+1}$. Otherwise, if the product of a vector in $V_n$ is bigger than α, the vector will be preserved. After it is confirm that each belief has bigger maximum product than the product in $V_n$, the control policy of this round is done.

It is obvious that the number and quality of sample beliefs is one of the important effect on control policy. A beliefs set with many duplicate and bias belief will lead to a low quality of control policy.

## Method

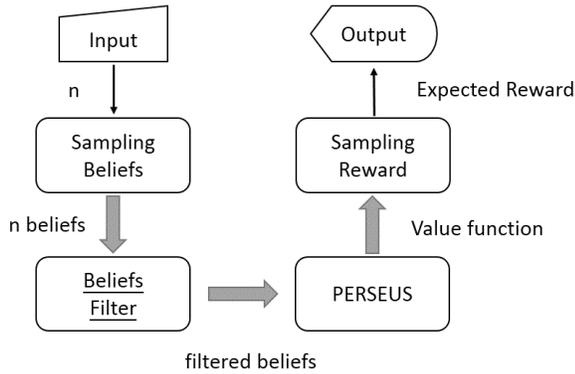

Figure 9: the flow chart of the evaluation process in this study.

The core of the method of I propose in this study is to add an extra stage, Beliefs Filter, in the original point-based approximate POMDP method. Figure 9 represents the operational flow chart of the method. Initially, a number, n, serves as the input of the sampling beliefs and the function will generate numbers of beliefs. Then a "Beliefs Filter" function, the core of the proposed method, filters out redundant approximate beliefs. The filtered beliefs are sent to "PERSEUS" for finding optimal policy. Then the next stage "Sampling Reward" simulates situations for evaluating the value function (i.e., control policy.) I employed "Hallway2" model as the benchmark function on this stage.

Algorithm 2: isSimilar function
function isSimilar(A,B, threshold)
INPUT: A,B = two vectors which going to be
          compared each other.
          threshold = the minimum difference for accept
          that A and B are similar.
OUTPUT: TRUE for A and B are similar, otherwise
          FALSE.
return ( max(abs(A-B))< threshold)

The function of the "beliefs filter" stage is to reduce the complexity of the repertoire of beliefs by eliminate similar beliefs. Specifically, this function carries out pairwise comparisons across all beliefs. The most important algorithm in "beliefs filter" stage is detailed in Algorithm 2 which calculates the distance between "A" and "B" vectors. If the distance is less than a "threshold", one of the two vectors can be eliminated. The value of threshold can be set between 0 and 1. The larger threshold, the more belief vectors will be eliminated by the "Beliefs Filter" function. The evaluation of PERSEUS with and without "Beliefs Filter" was carried out by "Sampling Reward" module (Algorithm 3). The starting states are all the states in POMDP model to the exclusion of goal states. In hallway2 problem, there are 88 non-goal states, which leads to 88 loops.

The input, value functions VF, is a set of vectors with corresponding best action. Each element of reward vector R stores the score of each trial in "SamplingRewards" (Algorithm 3). Line 11 in Algorithm 3 finds the best action for vector v such that the product of b on v is maximal. Line 14 determines the next belief by the traditional Bassian rule in POMDP(Russell and Norvig, 2009). Line 15 accumulates the value of each element in the reward vector R for each

---

Algorithm 3: testing the control quality of a POMDP solution

---
function sampleRewards
% Input VF: value function
% Output mean(R): Expected Reward
% variable R: Reward Vector
01: Trial_Per_Start = 10
02: Number_Of_States = 88
03: b0 = a uniform Belief
05: start = a vector which include all of the non-goal
          states number.
07: s0 = start(1)
08: step = 0
09: round = 1
10: while round ≤ (Number_Of_States *
                  Trial_Per_Start)
11:   $a = \text{action}(\arg\max_{v \in VF}(b * v))$
12:   draw s1 from P(s1|s0,a)
13:   draw o from P(o|s1,a)
14:   b = BassianRuleInPOMDP(b,a,o)
15:   R(round) = R(round)+
              RewardTable(s1,s0,a)*0.95^step
16:   step= step+1
17:   s0 = s1
18:   if s0 ∈ {Goal states}
19:     b = a uniform belief
20:     round = round+1
21:     s0 = start( ceil(round/ Trial_Per_Start) )
22:     step = 0
23:   end
24: end
25: return mean( R )

---

round by adding up the discounted reward scores of each step. The goal reward will be degraded significantly if it takes too many steps for reaching the goal. Line 18 is to check the goal state is reached, and, if has reached, reset to next round. Line 21, because each starting state is tried "Trial_Per_Start" times, the number of round divides by "Trial_Per_Start" and the ceiling number is a starting state. Finally, the algorithm returned mean of vector R, which returned an average score of all rounds.

## Evaluation

Table 1 : the result is from comparing original PERSEUS method with a method which have extra belief filtering stage. The number with underline is what I prefer to.

|  | Original | Beliefs Filter .01 |
|---|---|---|
| Sampling Beliefs | 3212 | 3212 |
| Threshold | N/A | 0.01 |
| Convergence | <u>0.0001</u> | 0.001 |
| Control Policy | 1031 | <u>1376</u> |
| Iteration | 142 | 95 |
| Expected Reward | 0.3468 | <u>0.3545</u> |
| Training Time (sec) | 1602 | 1420 |
| Filter Time (sec) | N/A | 156 |
| Total Time (sec) | 1602 | <u>1576</u> |

The empirical result is shown in Table 1. The mainframe of the test environment in this study is as below.
- Memory: DDR3-1333 28GB with ECC
- CPU: Intel Xeon 1230 v2
- OS: Windows 7 64bits
- Matlab R2013a 64bits

The original method includes 3212 random sample beliefs, which compares with the method with beliefs filter that includes 3212 sample beliefs which filtered from ten thousand random sample beliefs. The threshold is a parameter which be manually set in Algorithm 2. In this evaluation, the threshold is set to 1%. The convergence is as (5). "VB" is a matrix in which each element represents that each value function multiply each sample belief. "n" represents the number of iteration.

$$\text{convergence} = \frac{\text{sum}(VB_{n+1})}{\text{sum}(VB_n)} - 1 \quad (5)$$

"Control Policy" represents the vector number of policy, and a larger number represent a better solution in a situation. The iteration is the value iteration for improving value functions. The expected reward is a final score of value functions, it is generated from Algorithm 3. The filter time is how long it take to filtering the original sample beliefs. The result shows that the beliefs filter group have better quality of control policy by better expected reward and larger control policy. The empirical result shows that an approximate point-based POMDP method with beliefs filter does improve the training result, and the difference will be more significant when the training time is longer. In another perspective, the training speed will be faster when the training result is remained.

## Discussion

To clearly explain evaluation result, two states POMDP model and two dimensional value functions are shown in Figure 10. This example came from famous Tiger problem. In Figure 10(a), seven sample beliefs are generated and used. Those beliefs bias to one side in belief space, and many beliefs are redundant. It results that the final optimal policy can only handle limited situations, and the training time of redundant beliefs are unnecessary. It should be noticed that there is no "open right" action in Figure 10(a) because there is no such beliefs that can lead to this action. In Figure 10(b), seven filtered beliefs which refined from one hundred sample beliefs are used. It results that the beliefs cover all of the situation, and the final optimal policy will get higher expected reward in a simulation, and that explain the reason of the result in the evaluation.

POMDP could be used to solve complex Sequential decision problems and is wildly used in so many domains. Because the performance of the traditional formulas of POMDPs is too poor to be of practical use, the point-based approximate method, PERSEUS, shows a feasible method which is used in this study as a fundamental algorithm and a control group method. In a point-based method, some sample beliefs may look similar. Based on that the approximate solutions are acceptable in the used of the approximate method, an approximate sampling belief can be filtered out naturally. This study presents an extra beliefs filter stage in the point-based approximate POMDP method for increasing result quality and performance. The Hallway2 problem, a classical POMDP problem, is used as a benchmark. The empirical result shows that an approximate point-based POMDP method with beliefs filter does improve the performance.

Some people may over-generalize the approximate filtering concept in sampling method. It was not said that the filtering method in this study can be used in any sampling algorithm. For example, to calculate π by Monte Carlo method, the result will be a disaster if the similar sampling points is filtered out.

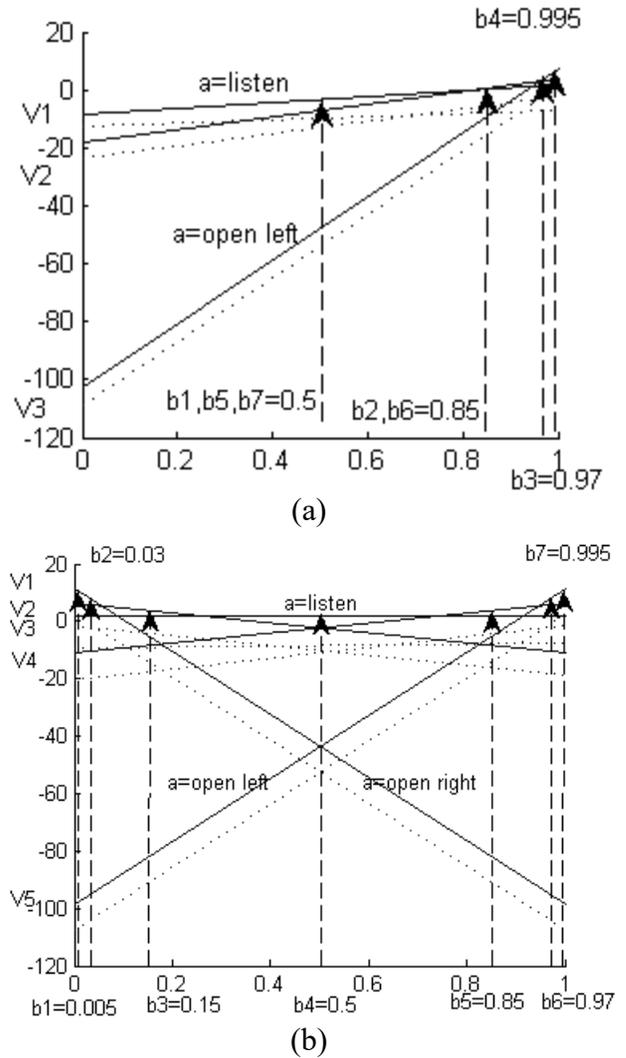

Figure 10: A two dimensional value functions of the backup stage in two states POMDP model. The solid lines are final policy, and "a" is action. There are two "listen" and one "open left". The dotted lines are results of previous backup stage. The dashed lines with arrow are beliefs. (a) Seven un-filtered beliefs (b) seven filtered beliefs which came from one hundred sample beliefs.